\documentclass{article}
\usepackage{spconf,amsmath,graphicx}
\usepackage{bbding}
\usepackage{booktabs}
\usepackage{colortbl}  
\usepackage{xcolor}
\usepackage{array}
\usepackage{amsmath,amsthm,amsfonts,amssymb,bm}
\usepackage{booktabs}
\usepackage{enumitem}
\usepackage{algorithm}
\usepackage{algorithmic}

\title{CSMPQ: Class Separability Based Mixed-Precision Quantization}
\name{Mingkai Wang$^{1}$ \qquad Taisong Jin$^{1}$ \qquad Miaohui Zhang$^{2}$ \qquad Zhengtao Yu$^{3,4}$}

\address{$^{1}$ Media Analytics and Computing Lab, Department of Computer Science and Technology,\\ School of Informatics,  Xiamen University, 361005, China. $^{2}$ Institute of Energy Research, \\Jiangxi Academy of Sciences.  $^{3}$ Kunming University of Science and Technology. \\$^{4}$ Yunnan Key Laboratory of Artificial Intelligence}

\begin{document}
%
\maketitle
\begin{abstract}
Mixed-precision quantization has received increasing attention for its capability of reducing the computational burden and speeding up the inference time. Existing methods usually focus on the sensitivity of different network layers, which requires a time-consuming search or training process. To this end, a novel mixed-precision quantization method, termed CSMPQ, is proposed. Specifically, the TF-IDF metric that is widely used in natural language processing (NLP) is introduced to measure the class separability of layer-wise feature maps. Furthermore, a linear programming problem is designed to derive the optimal bit configuration for each layer. Without any iterative process, the proposed CSMPQ achieves better compression trade-offs than the state-of-the-art quantization methods. Specifically, CSMPQ achieves 73.03$\%$ Top-1 acc on ResNet-18 with only 59G BOPs for QAT, and 71.30$\%$ top-1 acc with only 1.5Mb on MobileNetV2 for PTQ.
\end{abstract}
\section{Introduction}
\label{sec:intro}

Network quantization, which maps the single precision floating point weights or activations of the neural networks to lower bits for compression and acceleration, has received much attention of the researchers. The traditional quantization approaches ~\cite{zhou2017incremental,esser2019learned,nagel2019data,lee2021network} use the same low-bit for all network layers, which may cause the significant accuracy degradation. To address this drawback, mixed-precision quantization has recently been proposed, which can achieve the better trade-off between compression ratio and accuracy of the neural network. Currently, the representative mixed-precision quantization methods are as follows: HMQ~\cite{habi2020hmq} makes the bit-width and threshold differentiable using the Gumbel-Softmax estimator. HAWQ~\cite{dong2019hawq} leverages the eigenvalue of the Hessian matrix of weights for bit allocation. Network architecture search ~\cite{wu2018mixed,yu2020search} or reinforce learning ~\cite{elthakeb2018releq,wang2019haq} are used to search the optimal bit-width. 

Although mixed-precision quantization achieves promising learning performance in real applications, it still needs the numerous data and computational burdens. Specifically, the search space for a mixed-precision quantization is exponential to the number of the layers. Thus, it is intractable for the existing neural networks to handle the large-scale data, which limits the further performance enhancement of the mixed-precision quantization methods.

Inspired by the recent advances on network quantization, we propose a novel mixed-precision quantization method, termed CSMPQ. To measure the class separability of layer-wise feature maps, the proposed method introduces the TF-IDF metric in natural language processing to network quantization. Based on the derived layer-wise class separability via the TF-IDF metric, we design a linear programming problem to derive the optimal bit-width. In this way, the proposed method can allocate fewer bits to layers with lower class separability and vice versa. Without any iterative process, the proposed method can derive the optimal layer-wise bit-width in a few GPU seconds.

The main contributions of this work are three-fold: 

{(1)} We introduce the class separability of layer-wise feature maps to search for optimal bit-width. To our knowledge, the proposed method is the first attempt to apply the class separability of the layers to network quantization.

{(2)} We propose to leverage the TF-IDF metric that is widely used in NLP to measure the class separability of layer-wise feature maps.

{(3)} The extensive experiments demonstrate that the proposed method can provide the SOTA quantization performance and compression rate. The search process can be finished within 1 minute on a single 1080Ti GPU. 


\begin{figure*}[t]
	\centering
	\includegraphics[width=1.0\linewidth]{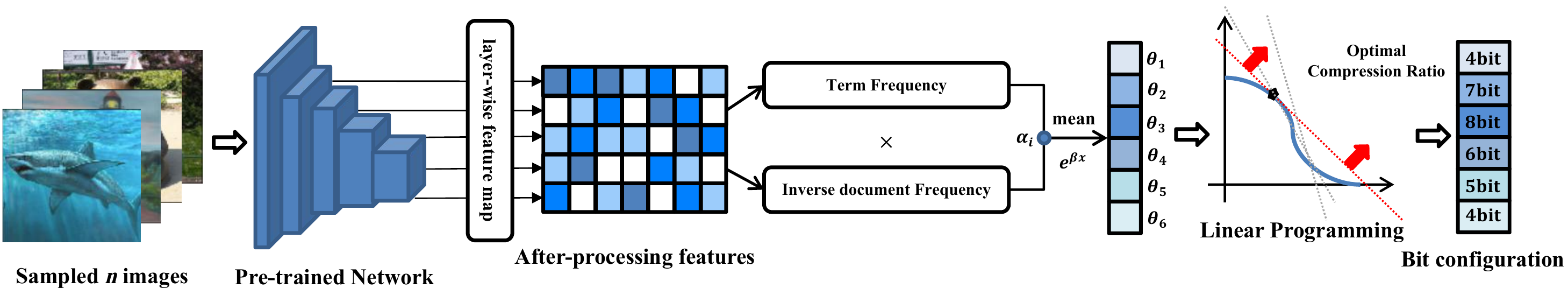}
	\caption{The overview of the proposed method. The \textit{n} images (\textit{n} = 32 in our experiments) are sampled to the pre-trained network, resulting in the layer-wise feature maps. And then, the derived feature maps are fed into an average pooling layer to reduce the dimension. Furthermore, based on the proposed TF-IDF metric, the layer-wise class separability scores are calculated to transform the class separability into layer-wise importance. In this way, the layer-wise importance and compression rate is formulated as a linear programming problem to drive the optimal bit-width configuration.}
	\label{fig:framework}
\end{figure*}

\section{Related Work}
\label{sec:format}

\subsection{Network Quantization} Existing network quantization can be roughly divided into two categories: quantization-aware training (QAT) and post-training quantization (PTQ). QAT ~\cite{zhou2017incremental,dong2019hawq} reduces the significant performance degradation by retraining. However, QAT is computationally expensive because of the fine-tuning process.  PTQ~\cite{cai2020zeroq,li2021brecq,nagel2019data} directly quantizes neural network models without fine-tuning. Mixed-precision quantization assigns the different bit-widths to the network layers across the model. Reinforcement learning ~\cite{elthakeb2018releq,wang2019haq} and network architecture search~\cite{wu2018mixed,yu2020search} are used to determine bit-widths. The existing methods usually require a lot of computation resources. Recently, second-order gradient information via the Hessian matrix was used to determine bit-widths ~\cite{dong2019hawq}. However, calculating the Hessian information of neural network is still time-consuming.

\subsection{TF-IDF} TF-IDF is widely-used in information retrieval and text mining. TF-IDF assesses the importance of a word to a document set or a document to a corpus. The importance of a word increases proportionally to the number of times a word appears in the document, but decreases inversely to the frequency a word appears in the corpus. 

TF-IDF for word \textit{t} in document \textit{d} from the document collection $D=\left\{d_{1}, \cdots, d_{j}\right\}$ is calculated as follows:
\begin{equation}
\operatorname{TF-IDF}_{\mathrm{i}, \mathrm{j}}=\mathrm{TF}_{\mathrm{i}, \mathrm{j}} \times \mathrm{IDF}_{\mathrm{i}},
\end{equation}%
where ${\displaystyle n_{i,j}}$ is the number of occurrences of a word in the document ${\displaystyle d_{j}}$, and the denominator is the sum of occurrences of all the words in the document ${\displaystyle d_{j}}$. 
TF refers to the frequency with which a given word appears in the document. For a word ${\displaystyle d_{j}}$ in a document, its importance can be calculated as 
\begin{equation}
\mathrm{TF_{i,j}} = \frac{n_{i,j}}{\sum_k n_{k,j}}.
\end{equation}%
IDF is a measure of the importance of a word. The IDF of a word can be obtained by dividing the total number of documents by the number of documents containing the word, and then taking the base 10 logarithm of the quotient which is defined as: 
\begin{equation}
\mathrm{IDF_{i}} = \lg \frac{|D|}{|\{j: t_{i} \in d_{j}\}|},
\end{equation}%
where ${n_{i,j}}$ is the number of occurrences of the word ${t_i}$ in the document ${d_j}$, and ${|D|}$ is the total number of documents in the corpus, $|\{j:t_{{i}}\in d_{{j}}\}|$ is the number of documents containing the term ${\displaystyle t_{i}}$. 
To avoid the denominator from being zero, $1+|\{j:t_{{i}}\in d_{{j}}\}|$ is used as the denominator.

High word frequency within a document, and low document frequency of that word in the whole corpus, result in a higher TF-IDF score. Thus, TF-IDF tends to filter out the common words and keep the important words.

\section{Methodology}
\label{sec:pagestyle}
  
\subsection{Pre-processing} 
Given a pre-trained model with $L$ network layers, $n$ images are sampled to obtain the feature map of each layer $\boldsymbol{X}^1, \boldsymbol{X}^2, \cdots, \boldsymbol{X}^L$. Then, the feature map $\boldsymbol{X}_j \in \mathbb{R}^{c^l \times h_{\text{out}}^l \times w_{\text{out}}^l \times {\text{1}} }$ of \textit{j}-th class are fed into an average pooling layer to reduce the feature dimension:
\begin{equation}\label{eq:pre_process}
{A}_{j}=\frac{\sum^{h_{\text {out }}} \sum^{w_{\text {out }}} \boldsymbol{X}_j}{h_{\text {out }} \times w_{\text {out }}},
\end{equation}%
where ${A}_{j} \in \mathbb{R}^{c_{\text {out }}}$ is the feature of each output channel after dimensionality reduction. We compose the features across different classes of the certain layer as ${A}=\left\{{A}_{1}, \cdots, {A}_{j}\right\} \in \mathbb{R}^{c_{\text {out }} \times j}$. After pre-processing, there are $c_{\text {out }}$ features in each layer for an image.\\
\textbf{Transforming the features into words.} The original TF-IDF metric used in NLP is designed to measure the importance of words in a document. It is unsuitable to directly leverage TF-IDF to measure the class separability of different layers. Thus, it is crucial to determining which features need to be converted into the words.

To derive the discriminative features for measuring the class separability of network layer, we would like to choose the features whose feature values deviates from the mean value of the features to be more sensitive and have a stronger representation capability. Thus, when a feature deviates from its mean value by a certain value, the corresponding feature is chosen as a word for the further computation. 
\begin{equation}
N_{i}=\left\{s_{j} \in S: abs({A}_{i, j}-\bar{{A}_{j}}>=\operatorname{std}\left({A}_{j}\right))\right\},
\end{equation}%
where \textit{S} is a set of the images. For the $j$-th image, ${A}_{i,j}$ is the $i$-th element of the feature map and $\bar{{A}_{j}}$ denotes the mean of the feature values. In this way, the suitable features are chosen as the words for computing the metric. \\
\subsection{The TF-IDF For Network Quantization} 

After transforming the features into the words, we formulate the TF of the $i$-th feature of the $j$-th image as
\begin{equation}\label{eq:new_tf}
\mathrm{TF}_{i, j}^{*}=\frac{{A}_{i, j}^{l}\times \operatorname{mask}\left({A}_{i, j}^{l}\in N_{i}\right) }{\sum_{k=0}^{c_{\text {out }}} {A}_{k, j}^{l}},
\end{equation}%
where ${A}_{i,j}^{l}$ represents the $i$-th element of the feature map in the $l$-th layer of the $j$-th image.

To make the features more discriminative, we use a mask to preserve the suitable features (words).
In this way, the defined TF can reflect the importance of different features in the $l$-th feature map. We define the IDF of the $i$-th feature as
\begin{equation}\label{eq:new_idf}
\mathrm{IDF}_{i}^{*}=\log \frac{1+|S|}{1+\left|N_{i}\right|},
\end{equation}%
where $|S|$ is the number of the small batch of the images, and $|N_{i}|$ is the number of the features that deviate from its mean value by a threshold. In this way, we derive the TF-IDF score of each feature in the $l$-th by multiplying improved TF with IDF as
\begin{equation}\label{eq:new_tfidf}
\mathrm{TF-IDF^{*}_{i, j}={TF}_{i, j}^{*} * {IDF}_{i}^{*}}.
\end{equation}%

We further use such TF-IDF of the feature to measure the importance of a network layer. Considering that if a layer has the more features with a higher TF-IDF score, then the corresponding class separability is more likely to be strong. Thus, the class separability of $l$-th layer can be defined as
\begin{equation}\label{eq:class_separability}
\mathrm{}
\alpha_{l}=\frac{{\sum_{k=0}^{c_{\text {out }}}\sum_{j}\mathrm{{TF-IDF}_{k, j}^{l}}}}{\left|N_{i}\right|\times{c_{\text {out }}}}, s_{j} \in S^{*}.
\end{equation}%

\begin{algorithm}[tb]
\caption{: CSMPQ}
\label{alg:csmpq}
\textbf{Input}: Pre-trained model \textit{M}, sampled \textit{n} i.i.d images \textit{D}.\\
\textbf{Output}: Optimal bit configuration \textbf{b} for each layer.

\begin{algorithmic}[1] 
\STATE Input D into M to obtain feature maps $\left\{{X}_{1}, \cdots, {X}_{L}\right\}$;
\FOR{${X}_{k}={X}_{1}, \cdots, {X}_{L}$}
\STATE Pre-process the feature maps of the $k$-th layer from ${X}_{1}, \cdots, {X}_{L}$ to ${A}_{1}, \cdots, {A}_{L}$ by Eq.~\ref{eq:pre_process};
\STATE Calculate the term frequency $\mathrm{TF}_{i, j}^{*}$ of each feature and the inverse document frequency $\mathrm{IDF}_{i}^{*}$ by Eq.~\ref{eq:new_tf} and Eq.~\ref{eq:new_idf};
\STATE Calculate the $\mathrm{TF-IDF^{*}_{i, j}}$ of each feature by Eq.~\ref{eq:new_tfidf};
\STATE Calculate the layer-wise class separability $\alpha_{k}$ by Eq.~\ref{eq:class_separability};
\STATE Obtain layer-wise importance $\theta_{k}$ by Eq.~\ref{eq:importance};
\ENDFOR
\STATE Solve the linear programming problem in Eq.~\ref{eq:objective} and Eq.~\ref{eq:constraint} to derive the optimal bit configuration \textbf{b};
\STATE \textbf{return} \textbf{b}
\end{algorithmic}
\end{algorithm}

\subsection{Mixed Precision Quantization} Given a pre-trained neural network, we use $\theta_{i}$ to denote the importance of $i$-th layer after obtaining the layer-wise class separability. The lower class separability $\alpha_{i}$ means the lower importance $\theta_{i}$ and vice versa. However, the class separability between each layer may vary dramatically. Thus, we leverage the monotonically increasing function $e^{x}$ to control the relative importance:
\begin{equation}\label{eq:importance}
\theta_{i}=e^{\beta \alpha_{i}},
\end{equation}%
where $\beta$ is a hyper-parameter that controls the relative importance to balance the bit-width between the different network layers. With the layer-wise importance, we then define a linear programming problem to maximize the global importance as follows:
\begin{gather}
    \text{Objective:}\label{eq:objective} \max _{\textbf{b}} \sum_{i=1}^{L}{b_{i}}\times{\theta_{i}},\\
    \text{Constraints:}\label{eq:constraint} \sum_{i}^{L} M{\left(b_{i}\right)} \leq \mathcal{T},
\end{gather}%
where $M{\left(b_{i}\right)}$ is the model size of the $i$-th layer when it is quantized to $b_{i}$-bit and $\mathcal{T}$ represents the model size.
Maximizing the objective function means assigning more bit-widths to the network layers with higher class separability, which implicitly maximizes the network’s representation capability. We use a scientific computation library SciPy~\cite{2020SciPy-NMeth} to solved this linear programming problem which only requires a few seconds on a single CPU.
Furthermore, the proposed method can serve as an auxiliary tool to search for optimal bit-width so that it can be easily combined with other quantization methods (including QAT and PTQ). The proposed algorithm is listed in Alg.~\ref{alg:csmpq}.


\section{Experiments}
\label{sec:typestyle}

In this section, we conduct the experiments on ImageNet to evaluate the effectiveness of the proposed CSMPQ. We first compare CSMPQ with the widely-used QAT methods. Then, we combine CSMPQ with a SOTA PTQ method BRECQ~\cite{li2021brecq} to further improve the accuracy at various compression rates.


\subsection{Mixed Precision}
We randomly sample 32 training data for all the models to obtain the layer-wise feature maps. We fix the first and last layer bit at 8 bit following the previous works. The search space of QAT is 4$ \sim $8 bit and that of PTQ is 2$ \sim $4 bit. The implementation for QAT is based on HAWQ-V3~\cite{yao2021hawq} and that for PTQ is based on BRECQ~\cite{li2021brecq}. In our experiment, only the bits for weight are quantized using mixed-precision, and the bits for activation are fixed. The whole search process 
only needs one forward pass which only costs about 30 seconds on a single 1080Ti GPU while other methods~\cite{dong2019hawq,yang2020fracbits,li2021brecq} need hundred of iterations.

\subsection{Quantization-Aware Training} 

We first conduct experiments of QAT on ImageNet-1k dataset. 
Two ResNet models with the different depths, namely ResNet-18 and ResNet-50 are chosen for the experiments. The experimental results are listed in Tabs.~\ref{tab:qat_res18} and ~\ref{tab:qat_res50}. 

As shown in Tabs.~\ref{tab:qat_res18} and ~\ref{tab:qat_res50}, compared to the SOTA QAT methods, the proposed CSMPQ has the better compression-accuracy trade-off. (1) For ResNet-18, CSMPQ achieves 73.03$\%$ Top-1 acc with only 59G BOPs and 6.7Mb when the activation is set to 6-bit.  Compared to HAWQ-V3, the proposed CSMPQ achieves 2.81$\%$ higher Top-1 acc while achieving 13G BOPs reduction. When the activation is set to 8-bit, CSMPQ achieves 73.16$\%$ Top-1 acc with only 79G BOPs. (2) For ResNet-50, CSMPQ achieves the Top-1 acc of 76.62$\%$ with only 143G Bops and 16.0Mb. The promising performance validates the effectiveness of CSMPQ.

\begin{table}[!t]
\centering
\resizebox{1.0\linewidth}{!}{
\begin{tabular}{@{}lccccc@{}}
\toprule
\textbf{Method} & \textbf{W bit} & \textbf{A bit} & \textbf{Model Size (MB)} & \textbf{BOPs (G)} &  \textbf{Top-1 (\%)}\\ \midrule
Baseline & 32 & 32 & 44.6 & 1858 & 73.09\\ \midrule
RVQuant~\cite{park2018value} & 8 & 8 & 11.1 & 116 & 70.01\\ 
HAWQ-V3~\cite{yao2021hawq} & 8 & 8 & 11.1 & 116 & 71.56\\ 
\rowcolor{gray!30}\textbf{CSMPQ} & \textbf{mixed} & 8 & \textbf{6.7} & \textbf{79} & \textbf{73.16}\\ \midrule
PACT~\cite{choi2018pact} & 5 & 5 & 7.2 & 74 & 69.80\\ 
LQ-Nets~\cite{zhang2018lq} & 4 & 32 & 5.8 & 225 & 70.00\\ 
HAWQ-V3~\cite{yao2021hawq} & mixed & mixed & 6.7 & 72 & 70.22\\ 
\rowcolor{gray!30}\textbf{CSMPQ} & \textbf{mixed} & 6 & \textbf{6.7} & \textbf{59} & \textbf{73.03}\\ \midrule
\end{tabular}
}
\caption{QAT experiments on ImageNet with ResNet-18.}
\label{tab:qat_res18}
\end{table}

\begin{table}[!t]
\centering
\resizebox{1.0\linewidth}{!}{
\begin{tabular}{@{}lccccc@{}}
\toprule
\textbf{Method} & \textbf{W bit} & \textbf{A bit} & \textbf{Model Size (MB)} & \textbf{BOPs (G)} &
\textbf{Top-1 (\%)} \\ \midrule
Baseline & 32 & 32 & 97.8 & 3951 & 77.72 \\ \midrule
PACT~\cite{choi2018pact} & 5 & 5 & 16.0 & 133 & 76.70 \\
LQ-Nets~\cite{zhang2018lq} & 4 & 32 & 13.1 & 486 & 76.40 \\
RVQuant~\cite{park2018value} & 5 & 5 & 16.0 & 101 & 75.60 \\ 
HAQ~\cite{wang2019haq} & mixed & 32 & 9.62 & 520 & 75.48 \\ 
OneBitwidth~\cite{chin2020one} & mixed & 8 & 12.3 & 494 & 76.70 \\ 
HAWQ-V3~\cite{yao2021hawq} & mixed & mixed & 18.7 & 154 & 75.39 \\ 
\rowcolor{gray!30}\textbf{CSMPQ} & \textbf{mixed} & 5 & \textbf{16.0} & \textbf{143} & \textbf{76.62} \\
\midrule
\end{tabular}
}
\caption{QAT experiments on ImageNet with ResNet-50.}
\label{tab:qat_res50}
\end{table}

\begin{figure} [tb]
\centering
\includegraphics[width=1.0\linewidth]{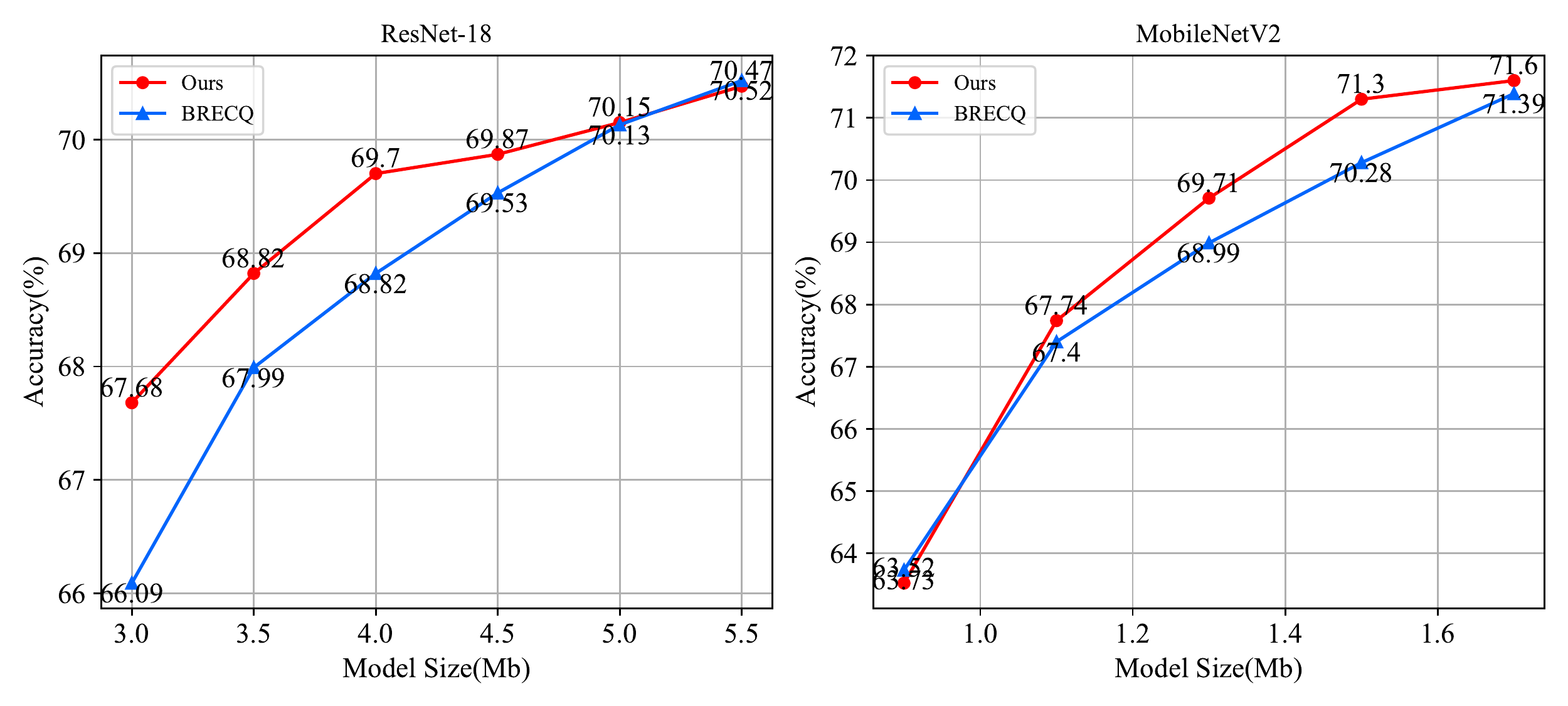}
\caption{PTQ experiment of CSMPQ combined with BRECQ.}
\label{fig:ptq}
\end{figure}

\subsection{Post-Training Quantization}
CSMPQ can combine with the other SOTA PTQ methods to further improve the performance. BRECQ~\cite{li2021brecq} performs the block reconstruction to reduce quantization errors. The experimental results are listed in Tabs .~\ref{tab:ptq_res18} and ~\ref{tab:ptq_mbv2}. Fig.~\ref{fig:ptq} shows the accuracy of BRECQ under the different model sizes.

As shown in Tabs. 3 and 4, for ResNet-18, CSMPQ combine with BRECQ outperforms the basic BRECQ by 0.88$\%$ Top-1 accuracy under a model size of 4.0Mb; For MobileNet-V2, CSMPQ achieves a higher accuracy of (71.30$\%$ vs 69.90$\%$) while keeping a lower model size (1.5Mb vs 1.8Mb). As clearly seen in Fig.~\ref{fig:ptq}, CSMPQ improves the accuracy of BRECQ under the different model sizes. 

\begin{table}[]
\centering
\resizebox{0.9\linewidth}{!}{ 
\begin{tabular}{@{}lcccc@{}}
\toprule
\textbf{Method} & \textbf{W bit} & \textbf{A bit} & \textbf{Model Size (Mb)} & \textbf{Top-1 (\%)}\\ \midrule
Baseline & 32 & 32 & 44.6 & 71.08\\ 
FracBits-PACT & mixed & mixed & 4.5 & 69.10\\ 
\rowcolor{gray!30}\textbf{CSMPQ+BRECQ} & mixed & 4 & 4.5 & 69.00\\ 
\rowcolor{gray!30}\textbf{CSMPQ+BRECQ} & \textbf{mixed} & \textbf{8} & \textbf{4.5} & \textbf{69.87}\\ \midrule
PACT~\cite{choi2018pact} & 4 & 4 & 5.81 & 69.20\\ 
HAWQ-V3~\cite{yao2021hawq} & 4 & 4 & 5.81 & 68.45\\ 
FracBits-PACT~\cite{yang2020fracbits} & mixed & mixed & 5.81 & \textbf{69.70}\\ 
\rowcolor{gray!30}\textbf{CSMPQ+BRECQ} & \textbf{mixed} & \textbf{4} & \textbf{5.5} & 69.43\\ \midrule
BRECQ~\cite{li2021brecq} & mixed & 8 & 4.0 & \textbf{68.82}\\ 
\rowcolor{gray!30}\textbf{CSMPQ+BRECQ} & \textbf{mixed} & \textbf{8} & \textbf{4.0} & \textbf{69.70}\\ \midrule
\end{tabular}
}
\caption{PTQ experiments on ImageNet with ResNet-18.}
\label{tab:ptq_res18}
\end{table}

\begin{table}[!t]
    \centering
    \resizebox{0.9\linewidth}{!}{
        \begin{tabular}{@{}lcccc@{}}
        \toprule
        \textbf{Method} & \textbf{W bit} & \textbf{A bit} & \textbf{Model Size (Mb)} & \textbf{Top-1 (\%)} \\ \midrule
        Baseline & 32 & 32 & 13.4 & 72.49\\ \midrule
        BRECQ~\cite{li2021brecq} & mixed & 8 & 1.3 & 68.99\\ 
        \rowcolor{gray!30}\textbf{CSMPQ+BRECQ} & mixed & \textbf{8} & \textbf{1.3} & \textbf{69.71}\\ \midrule
        FracBits~\cite{yang2020fracbits} & mixed & mixed & 1.84 & \textbf{69.90}\\ 
        BRECQ~\cite{li2021brecq} & mixed & 8 & 1.5 & 70.28\\ 
        \rowcolor{gray!30}\textbf{CSMPQ+BRECQ} & \textbf{mixed} & \textbf{8} & \textbf{1.5} & \textbf{71.30}\\  \midrule
        \end{tabular}
    }
    \caption{PTQ experiments on ImageNet with MobileNetV2.}
    \label{tab:ptq_mbv2}
\end{table}

\section{Conclusion}
\label{sec:majhead}

In this paper, we have proposed a novel mixed-precision quantization method (CSMPQ), which calculates the class separability of layer-wise feature maps using TF-IDF that is widely used in natural language processing. 
The whole search process costs only a few seconds on a single 1080Ti GPU. In both QAT and PTQ, CSMPQ can achieve the better compression trade-offs than the existing methods. 


\vfill\pagebreak

\bibliographystyle{IEEEbib}
\bibliography{strings}

\end{document}